\documentclass[10pt,journal,compsoc]{IEEEtran}
\usepackage{ulem}
\usepackage{times}
\usepackage{epsfig}
\usepackage{graphicx}
\usepackage{amsmath}
\usepackage{amssymb}
\usepackage{amsmath}
\DeclareMathOperator*{\argmin}{argmin}
\usepackage[pagebackref=true,breaklinks=true,letterpaper=true,colorlinks,bookmarks=false]{hyperref}
\usepackage{algorithm}
\usepackage{algorithmic}

\ifCLASSOPTIONcompsoc
  \usepackage[nocompress]{cite}
\else
  \usepackage{cite}
\fi

\ifCLASSINFOpdf
\else
\fi

\begin{document}
\title{GP-Aligner: Unsupervised Non-rigid Groupwise Point Set Registration Based On Optimized Group Latent Descriptor}
\author{Lingjing Wang, Xiang Li, Yi Fang
\IEEEcompsocitemizethanks{\IEEEcompsocthanksitem L.Wang is with MMVC Lab, the Department of Electrical Engineering, New York University Abu Dhabi, UAE, e-mail: lingjing.wang@nyu.edu. X.Li is with the MMVC Lab, New York University Abu Dhabi UAE, e-mail: xl1845@nyu.edu. Y.Fang is with MMVC Lab, Dept. of ECE, NYU Abu Dhabi, UAE and Dept. of ECE, NYU Tandon School of Engineering, USA, e-mail: yfang@nyu.edu.
\IEEEcompsocthanksitem Corresponding author: Yi Fang. Email: yfang@nyu.edu.}
\thanks{Manuscript received April 19, 2005; revised August 26, 2015.}}
\markboth{Journal of \LaTeX\ Class Files,~Vol.~14, No.~8, August~2015}%
{Shell \MakeLowercase{\textit{et al.}}: Bare Advanced Demo of IEEEtran.cls for IEEE Computer Society Journals}

\IEEEtitleabstractindextext{%
\begin{abstract}
In this paper, we propose a novel method named GP-Aligner to deal with the problem of non-rigid groupwise point set registration.  Compared to previous non-learning approaches, our proposed method gains competitive advantages by leveraging the power of deep neural networks to effectively and efficiently learn to align a large number of highly deformed 3D shapes with superior performance. Unlike most learning-based methods that use an explicit feature encoding network to extract the per-shape features and their correlations, our model leverages a model-free learnable latent descriptor to characterize the group relationship. More specifically, for a given group we first define an optimizable Group Latent Descriptor (GLD) to characterize the gruopwise relationship among a group of point sets. Each GLD is randomly initialized from a Gaussian distribution and then concatenated with the coordinates of each point of the associated point sets in the group. A neural network-based decoder is further constructed to predict the coherent drifts as the desired transformation from input groups of shapes to aligned groups of shapes. During the optimization process, GP-Aligner jointly updates all GLDs and weight parameters of the decoder network towards the minimization of an unsupervised groupwise alignment loss. After optimization, for each group our model coherently drives each point set towards a middle, common position (shape) without specifying one as the target. GP-Aligner does not require large-scale training data for network training and it can directly align groups of point sets in a one-stage optimization process. GP-Aligner shows both accuracy and computational efficiency improvement in comparison with state-of-the-art methods for groupwise point set registration. Moreover, GP-Aligner is shown great efficiency in aligning a large number of groups of real-world 3D shapes. 
\end{abstract}

}

\maketitle
\IEEEdisplaynontitleabstractindextext
\IEEEpeerreviewmaketitle
\ifCLASSOPTIONcompsoc
\IEEEraisesectionheading{\section{Introduction}\label{sec:introduction}}
\else
\section{Introduction}
\label{sec:introduction}
\fi

\IEEEPARstart{P}{oint} set registration is a fundamental computer vision task, which plays an important role in many applications such as image registration, object correspondence, large-scale 3D reconstruction, and so on \cite{myronenko2009image,ma2016non,maintz1998survey,wang2019deep}. Before the era of deep learning, traditional point set registration methods usually search the optimal geometric transformation by minimizing a pre-defined alignment loss between the transformed shape and target shape in an iterative process \cite{myronenko2007non,ma2014robust,ling2005deformation,viergever2016survey}. Traditional non-learning based methods can be divided into two branches: feature-based methods and intensity-based methods. Feature-based methods usually leverage hand-crafted features to match points in source and target point sets. For example, one of the most popular methods Iterative Closest Point (ICP) \cite{besl1992method} estimates the rigid transformation by finding a set of corresponding points. Other methods formulate a probability density distribution function from one point set and fit the other point set to this distribution to maximize the accumulative density likelihood defined as a similarity metric, e.g., CPD \cite{myronenko2009image}. Intensity-based methods tend to directly optimize the transformation matrix by minimizing a pre-defined metric between intensity patterns of source and target shapes. This category of methods is widely used in medical images/voxels registration \cite{sotiras2013deformable}. In recent years, as deep learning-based methods have achieved great success in various visual recognition tasks, researchers are increasingly interested in bringing in deep learning-based solutions to the field of point set registration. Thanks to the powerful feature learning/representation abilities, learning-based methods demonstrated high potential in this field \cite{liu2019flownet3d,balakrishnan2018unsupervised,rocco2017convolutional,chen2019arbicon}. 

In this paper, we focus on the task of aligning a group of point sets towards one common/middle shape (position). In many real-world applications, e.g., registration of a sequential of medical images \cite{che2019deep}, 3D mapping \cite{wang2020unsupervised, Ding_2019_CVPR} and so on, there's an urgent need for developing methods for registering a group of point sets/images. However, direct extension from pair-wise registration to groupwise registration is non-trivial. Recently, Che et al. \cite{che2019deep} propose a learning-based algorithm for the groupwise registration of multi-spectral fundus images. They enhance the model proposed in \cite{balakrishnan2018unsupervised} by dynamically computing the templates from a group of transformed input voxel images and compare each input voxel image in the group with the templates using a pre-defined loss \cite{balakrishnan2018unsupervised}. Some unsolved problems such as how to model the relationship among a group of point sets, a proper definition of the groupwise similarity measure, choose of a template of the group (in practice, the assumption of the existence of an actual template to match against is hard to make) are usually not considered in pair-wise registration task and make the groupwise registration problem even more challenging. 

In comparison to the pair-wise point sets registration task, only a few recent researches \cite{giraldo2017group,chen2010group,wang2006groupwise} have studied the problem of groupwise point set registration. In general, these methods fall into two categories. The first category of methods works by selecting a corresponding subset of input point sets as a reference to solve the desired transformations \cite{chui2003new,chui2004unsupervised}. These methods are less robust and usually have difficulties in dealing with various noise and outliers. The second category of methods \cite{giraldo2017group,chen2010group,wang2006groupwise} directly register the point sets from the population without detecting correspondence. For these methods, the main challenge is to define an efficient groupwise similarity-based cost function. Chen et al. \cite{chen2010group} introduce CDF-HC divergence to quantify the similarity among all Cumulative distribution function (CDF) estimated from each point set. Recently, \cite{giraldo2017group} provides a closed-form solution to optimize the previously defined cost function proposed in \cite{chen2010group} by using Renyi's second-order entropy and improved optimization efficiency. Nevertheless, these methods suffer from huge computation costs because they take the entire point sets as inputs. 

In comparison to classical optimization-based methods, deep learning-based methods have been proved to be more robust for feature learning and enjoy higher efficiency due to powerful GPU parallel computation. In addition to the challenges listed above, learning-based methods usually require a well-labeled dataset for supervised training but labeled datasets are usually unavailable for groupwise registration tasks. In contrast to the existing methods, we propose an unsupervised deep learning-based optimization algorithm for the task of groupwise point set registration. Our method can not only leverage the power of the neural network structure but also does not require any training process. The alignment of groups of shapes can be completed during a single optimization process. Moreover, our algorithm does not need a pre-defined middle shape but can atomically search the optimal one during the optimization process and aligns all the input groups with great efficiency. 

\begin{figure*}
\begin{center}
\includegraphics[width=17cm]{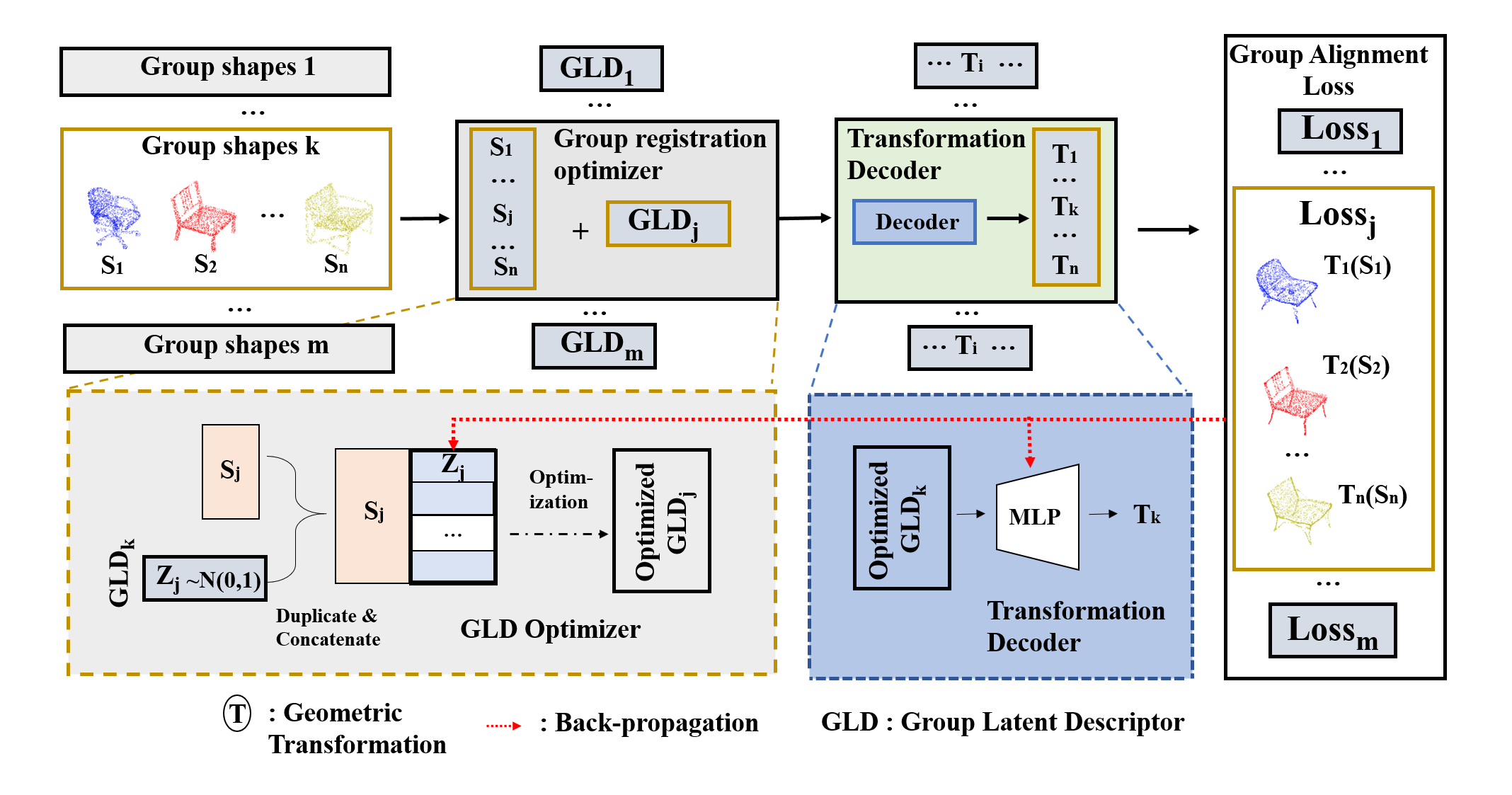}
\end{center}
\caption{Our pipeline. Our method contains three main components. The first component is a group registration optimizer where the GLDs are optimized from a set of randomly initialized vectors. The second component is the transformation decoder which decodes the GLDs to the desired transformation for each group. The third component is a groupwise alignment loss that measures the similarity among transformed shapes in each group. The communication route in red is the back-propagation route with which the groupwise alignment loss is back-propagated to update the GLDs and the transformation decoder. }
\label{main}
\end{figure*}
As shown in Figure \ref{main}, our proposed GP-Aligner framework mainly contains three components. In the first component, we define a set of optimizable Group Latent Descriptors (GLDs), one for each group, to characterize the global group features of each group and also the mutual information among a group of shapes. Each GLD is randomly initialized from a Gaussian distribution and optimized towards the minimization of a pre-defined alignment loss. In the second component, given each input shape and the associated optimized GLD, an MLP-based transformation decoder further decodes the GLD to a coherent drift field as the desired transformations from each of the input shape among the group to the target position. In the third component, we formulate a groupwise alignment loss to measure the similarity among all transformed point sets.

Our contribution is as follows:

\begin{itemize}
\item We introduce a novel deep neural network-based method for the task of groupwise point set registration. Our proposed model leverages the power of deep neural networks but can be optimized to align a group of point sets without training.\\

\item We introduce a set of Group Latent Descriptors (GLDs), one for each group, to eliminate the challenge of encoding the groupwise relationship among a group of point sets. The GLDs can be optimized along with the network parameters towards minimization of pre-defined groupwise alignment loss function.\\

\item Experimental results demonstrate the effectiveness and high efficiency of the proposed method for groupwise point set registration, especially for registration of a large number of groups at the same time.

\end{itemize}
\section{Related works}

\subsection{Pairwise registration}
The classical algorithm Iterative Closest Point (ICP) algorithm \cite{besl1992method} is one successful solution for rigid registration. To accommodate the deformation (e.g. morphing, articulation) between a pair of point sets, many efforts were spent on the development of algorithms for a non-rigid transformation. For non-rigid registration, classical previous methods can usually be divided into parametric and non-parametric by the target transformation. As a robust parametric method, the TPS-RSM algorithm was proposed by Chui and Rangarajan \cite{chui2000new} to estimate parameters of non-rigid transformation with a penalization on second-order derivatives. As a classical non-parametric method, coherence point drift (CPD) was proposed by Myronenko et al. \cite{myronenko2007non}, which successfully introduced a process of fitting Gaussian mixture likelihood to align the source point set with the target one. With the penalization term on the velocity field, the algorithm enforces the motion of the source point set to be coherent during the registration process. Another non-parametric vector field consensus algorithm was proposed by Ma et al. \cite{ma2014robust}. This algorithm is emphasized to be robust to outliers. For learning-based methods, Balakrishnan et al. \cite{balakrishnan2018unsupervised} proposed a voxel morph structure to align two volumetric 3D shapes/images using a U-net structure in an unsupervised way. Liu et al. \cite{liu2019flownet3d} proposed a supervised flow network to estimate end-to-end non-rigid drifts from source point set to target point set for their alignment. Wang et al. \cite{wang2019non, wang2019coherent} proposed unsupervised networks for learning pairwise non-rigid point set registration.  

\subsection{Groupwise registration}
Even though groupwise registration can be considered as an extension task from pair-wise registration and shares the same challenges as discussed in the previous session, the task of groupwise registration has its challenges as well. Most classical groupwise registration methods need two steps for this task: step one is to find a mean shape; and step two is to register all the shapes of the group towards the mean shape. Abtin et al. \cite{rasoulian2012group} proposed a groupwise registration technique that leverages soft correspondences between groups of point sets. By learning the shape variation of the group, the learned model of a mean shape can be further aligned towards all the shapes of the group after a transformation. Chui et al. \cite{chui2004unsupervised} leverages a joint clustering and matching algorithm for computation of a mean shape from multiple unlabelled shape samples. Then all the shapes of the group move towards the mean shape as an iterative bootstrap process. In comparison, our method does not require a separate two steps process for learning the middle shape and registration. Equipped with deep neural networks-based structure, the optimization process is naturally merged with the stochastic gradient algorithm which guides all the shapes to move toward the middle one. Chen et al. \cite{chen2010group}
introduced CDF-HC divergence to quantify the dissimilarity between estimated CDFs from individual point sets which further generalizes previous research on CDF-JS divergence introduced by \cite{wang2006groupwise}. To further improve the efficiency of the divergence algorithm, Luis et al. \cite{giraldo2017group} derived a closed-form formula for the analytic gradient for CDF-HC divergence, which is more efficient in comparison to the previous algorithm. In comparison, based on our deep neural network-based optimized feature we completely propose a novel different approach to this question. Our model is proved to be much more efficient than \cite{giraldo2017group}. Even though there are less learning-based researches introduced for groupwise point set registration, learning methods for aligning a group of medical images is being paid a lot of attention recently. Che et al. \cite{che2019deep} extended the pairwise voxelmorph model proposed by \cite{balakrishnan2018unsupervised} for groupwise registration for Multi-Spectral Images From Fundus Images. Siebert et al. \cite{siebert2020deep} proposed a deforming autoencoder for deep groupwise registration of MRI brain scans in an unsupervised learning setting. However, these methods all rely on encoding network to exact shape features and further modeling a groupwise relationship based on features. Exacting features from 3D voxel and point sets are quite challenging and the relationship between a group of images is even more difficult to model. Instead of encoding the features from input shapes, our model leverages an optimized descriptor for exacting the necessary information from the pack-propagation of the groupwise alignment loss. 

\section{Methods}

We introduce our approach in the following sections. First, we define the groupwise registration problem in section \ref{sc_problem}. In section \ref{sc_scr}, we introduce the shape registration optimizer. Section \ref{sc_trans_reg} illustrates the coherent flow field decoding process. In section \ref{sc_loss}, we provide the definition of our proposed groupwise similarity function. The optimization process is illustrated in section \ref{sc_optim}.

\subsection{Problem Statement}\label{sc_problem}

For groupwise registration task, given a dataset $\mathcal{D}$ of $M$ groups $\mathcal{S}_1,...,\mathcal{S}_M$ and for each group we have $K$ point sets ($K$ can be different for different groups). Assuming the existence of a MLP-based parametric function $g_{\theta}$ with weights (parameters) $\theta$, $\forall {S_k^{(i)}} \in \mathcal{S}_i$, the desired geometric transformation can be formulated as $T_k^{(i)}= g_{\theta}({S_k^{(i)}}, Z_i)$, where $Z_i \sim \mathcal{N}(0,0.01)$ is an optimizable Group Latent descriptor associated with shape ${S_k}$, such that the transformed point sets $\{ T_k^{(i)}({S_k^{(i)}}) | S_k^{(i)} \in \mathcal{S}_i \}_{(k=1,2,...,K)}$ can reach the minimal pre-defined groupwise similarity-based loss for all groups $\{\mathcal{S}_i\}_{i=1,...,M}$. The geometric transformation in this paper is represented by the coordinate drifts from each point in $S_k^{(i)}$ to the corresponding location in $T_k^{(i)}({S_k^{(i)}})$. A gradient descent based algorithm is used to update the weights parameters $\theta$ and latent code $\{Z_k\}_{k=1,...,M}$ towards the minimization of a pre-defined loss:
\begin{equation}
\begin{split}
\tilde{{\theta}}, \tilde{Z}_1,...,\tilde{Z}_M =\argmin_{\theta, Z_1,..,Z_M}[\mathbb{E}_{\{\mathcal{S}_i\}\sim \mathcal{D}}[\mathbb{E}_{\{S_j^{(i)}\}\sim \mathcal{S}_i}[\\
\mathcal{L}(g_{\theta}({S_1^{(i)}},Z_i),...,g_{\theta}({S_K^{(i)}}, Z_i)]]],
\end{split}
\end{equation}
where $\mathcal{L}$ represents a groupwise similarity loss function and $\tilde{{\theta}}, \tilde{Z}_i$ is the optimized latent descriptor for group ${\mathcal{S}_i}$. After the optimization process, the desired transformation $\{ \tilde{T}^{(i)}_k = g_{\tilde{\theta}}({S_k^{(i)}}, \tilde{Z}_i) \}_{k=1,...,K}$ can be regarded as the desired solution for aligning group $\mathcal{S}_i$. In practice, during the optimization process, we need to balance the penalization term on the smoothness and scale of the predicted drifts together with the alignment loss in function $\mathcal{L}$. 

\begin{algorithm*}
\caption{groupwise registration optimization process.}
\begin{algorithmic}[1]
\STATE Initialize the decoder parameters $\theta^{(0)}$ and GLDs $\{ Z_1^{(0)},...,Z_m^{(0)}\}$ and choose hyper-parameters
\WHILE{not convergence}
\FOR{ each ${\mathcal{S}_i} \in \mathcal{D}$}
\FOR{ each ${S_j^{i}} \in \mathcal{S}_i$}
\STATE Compute the flow drifts $\{{dx}\}_{x\in S_j^i}$ and transformed point set $T^{i}_{j(\theta^{(k)})}({S_j^i}, Z_i^{(k)})$ for each input shape by Eq. (\ref{eq33}) and Eq. (\ref{eq44})
\ENDFOR 
\ENDFOR 
\STATE Compute the groupwise alignment loss for groups $\mathcal{L}^{(k)}$ on all transformed shapes $\{\{T^i_{j(\theta^{(k)})}({S_j^i}, Z_i^{(k)})\}_{j={1,...,K}} \}_{i=1,...,M}$ by Eq. (\ref{eq55}) 
\STATE Compute the gradients $dZ_i^{(k)}$ and $d\theta_i^{(k)}$ w.r.t the loss function $\mathcal{L}$ and updates $\theta^{(k+1)}$ and $\{ Z_i^{(k+1)} \}_{i=1,...,K}$
\ENDWHILE
\label{algo2}
\end{algorithmic}
\end{algorithm*}

\subsection{Shape Registration Optimizer}\label{sc_scr}
For each input group, we define an optimizable group latent descriptor to characterize its group geometric feature. Usually, this feature should be able to characterize the relationship of the input shapes and guide the transformation prediction process. Unlike volumetric shapes which are regularized in ordered standard voxels, point set data contains sparse geometric coordinates that are not sorted in any way. Even though previous works (e.g., PointNet \cite{qi2017pointnet}) have provided an efficient way to extract shape features from unstructured point set and recent researches (e.g., \cite{rocco2017convolutional}) attempted to formulate the correlation between shapes, it is still very challenging to design a generalized feature encoder for this task with more than two shapes in a group for registration and the target middle shape of a group is not even known. Furthermore, encoder networks are rather specialized and inept of dealing with data of large differences.  

To eliminate the side effects of the explicit design of the feature encoding network and correlation module, we directly define an optimizable Group Latent Descriptor (GLD) to represent the group features for the groupwise point set registration task. The GLD is randomly initialized from a zero-mean Gaussian distribution and then it will be concatenated to each point of input shape to guide the transformation prediction process. The GLD is initialized independently for each group in the dataset. Even though GLD is not explicitly learned from input shapes, we make it optimizable from the groupwise alignment loss together with the decoder network parameters. In this way, our method cannot only avoid the challenging problem of explicit defining the relationship among a group of point sets but also leverage the deep neural networks-based structure in the optimization framework.

\subsection{Coherent Drifts Field Decoding Process}\label{sc_trans_reg}
We define the geometric transformation $T$ as the coordinate drifts from each point $x$ in ${S_k}$ to the corresponding location in $T_k({S_k})$,
\begin{equation}
\label{eq:31}
\begin{split}
T(x,v) = x + f(x)
\end{split}
\end{equation} 
where $f: \mathbb{R}^2 \to \mathbb{R}^2 /\mathbb{R}^3 \to \mathbb{R}^3$ is a ``point displacement'' function. The groupwise point set registration task requires us to determine this displacement function $v$ such that all the point sets in the population can be coherently moved towards a common middle point set of group. To ensure a good groupwise alignment, one would expect that a model can align all point sets in the group to the same middle shape. Besides that, it is also necessary for function $f$ to be a continuous and smooth function according to the Motion Coherent Theory (MCT) \cite{yuille1989mathematical}. Our deep neural networks-based decoder is well designed so that the desired function $f$ via our proposed MLP architecture can not only aligns the source and target point sets but also satisfies the continuous and smooth characteristics. Please refer \cite{chen2019arbicon} for further explanation.

For registration of a number of groups, given the above defined GLD for each group, the input of decoder is formulated by stacking the coordinates of each point in individual shape ${S_k^{(i)}}$ with the GLD, noted as $[x^{(i)}, {Z_i}], \forall x \in {S_k^{(i)}}$. Then we define a multi-layer perceptron (MLP) architecture for learning coherent drifts flow. This architecture includes successive MLP layers with the ReLU activation function, $\{g_i\}_{i=1,2,...,s}$, such that $g_i : \mathbb{R}^{v_{i}}\to \mathbb{R}^{v_{i+1}}$, where $v_{i}$ and $v_{i+1}$ are the dimensions of the layer inputs and outputs respectively, calculated as:

\begin{equation}
\begin{split}
{dx_k^{(i)}}=g_sg_{s-1}...g_1([{x_k^{(i)}},Z_i])
\end{split}
\label{eq33}
\end{equation}

\begin{equation}
\begin{split}
T_k^{(i)}({S_k^{(i)}})= \{{x_k^{(i)}}+ {dx_k^{(i)}}\}_{{x_k^{(i)}}\in {S_k^{(i)}}}
\end{split}
\label{eq44}
\end{equation}
where $T_k^{(i)}$ denotes the flow field as the transformation function for each shape ${S_k^{(i)}} \in \mathcal\bold{S}_i$. After generating all the transformed point sets from the group, we further define a groupwise alignment loss function to optimize the GLDs and decoder network parameters in the next section.
\begin{figure*}
\begin{center}
\includegraphics[width=14cm]{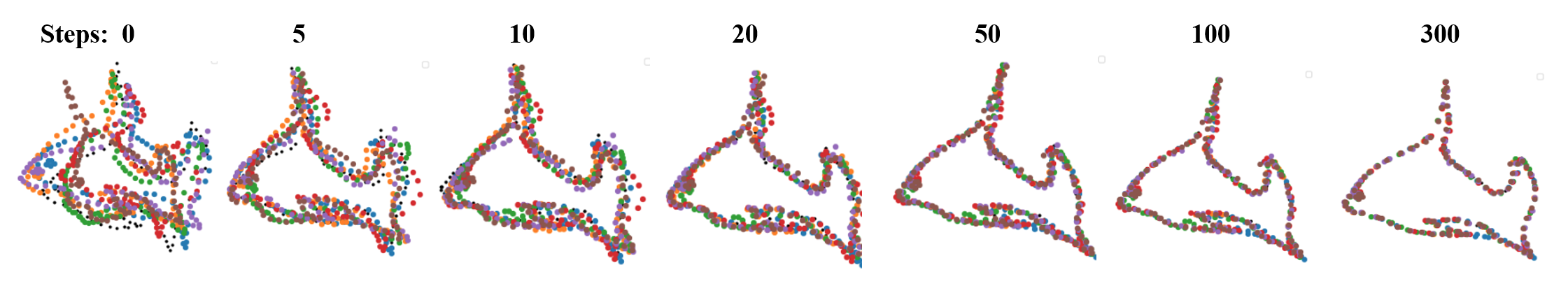}
\end{center}
\caption{Groupwise point set alignment process. The inputs include 7 non-rigid deformed shapes and the deformation level of inputs is 0.4.}
\label{steps}
\end{figure*}
\begin{figure*}
\begin{center}
\includegraphics[width=12cm]{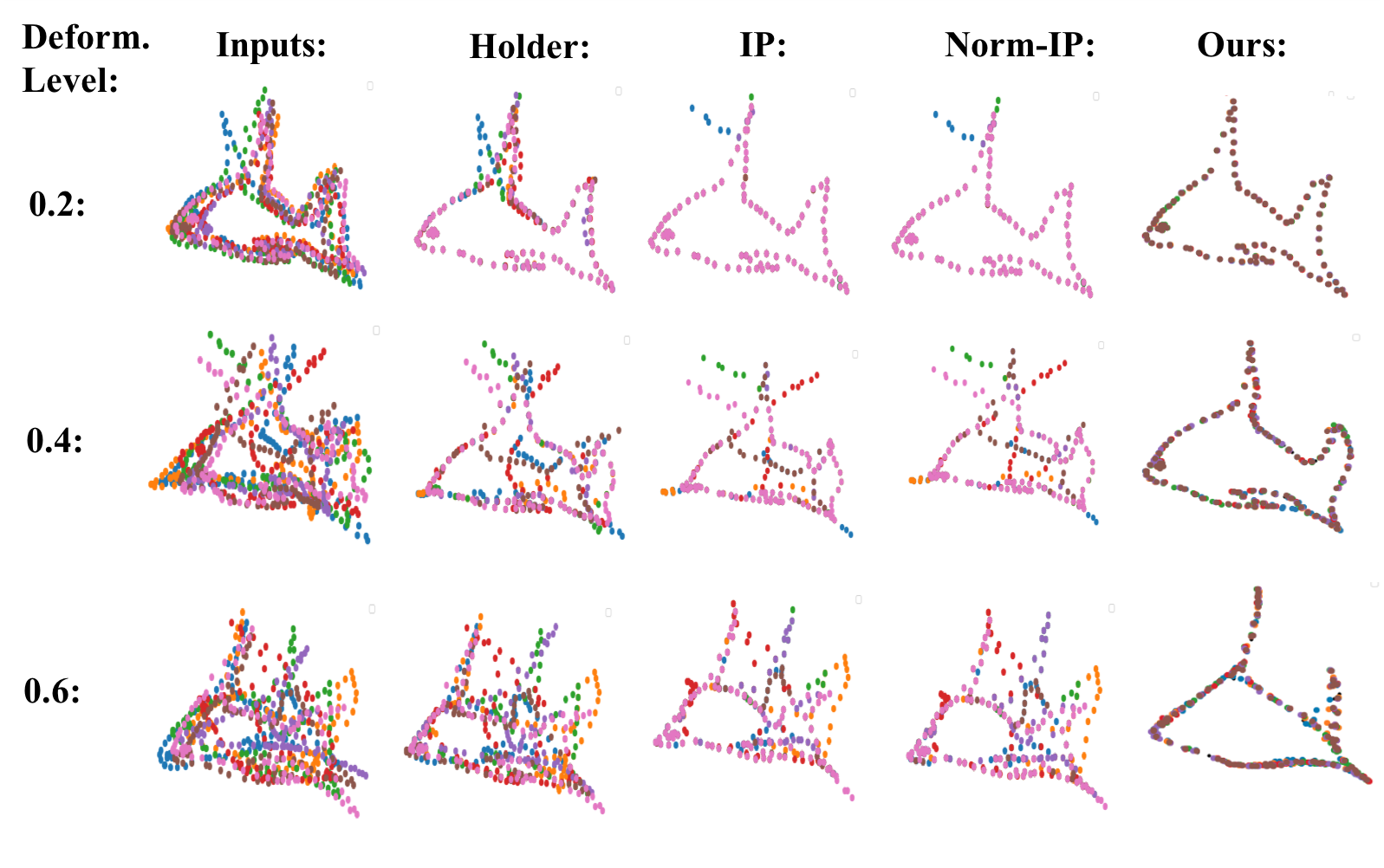}
\end{center}
\caption{The qualitative registration results for Fish shape at different deformation levels. Inputs include seven fish-shape point sets in different colors with various non-rigid deformation.}
\label{exp1}
\end{figure*}
\subsection{Loss function}\label{sc_loss}
In our unsupervised setting, we do not have the ground truth transformation for supervision and we do not assume correspondences between any pair of point sets. Therefore, a distance metric between two point sets, instead of the point-/pixel-wise loss is desired. In addition, a suitable metric should be differentiable and efficient to compute. In this paper, we adopt the Chamfer distance proposed in \cite{fan2017point} as our loss function. The Chamfer loss is a simple but effective alignment metric defined on two non-corresponding point sets. We formulate the groupwise Chamfer loss among a population of point sets $\mathcal{S}=\{{S_1},...,{S_k} \}$ as:

\begin{equation} 
\begin{split}L_{\text{Group-Chamfer}}(T(\mathcal{S}))&= \sum_{{S_i} \in \mathcal{S}} \sum_{{S_j} \neq {S_i}} L_{\text{Chamfer}}(T_i({S_i}),T_j({S_j}))
\end{split}
\label{eq5}
\end{equation}
, where the Chamfer loss between each pair $L_{\text{Chamfer}}$ is formulated as:
\begin{equation} 
\begin{split}L_{\text{Chamfer}}(X,Y)
&= \sum_{x\in X}\min_{y \in Y}||x-y||^2_2\\
&+ \sum_{y\in Y}\min_{x \in X}||x-y||^2_2
\end{split}
\end{equation}

We notice that we need to add a regularization term to regularize the scale of drifts (transformation) in practice and we have our regularized groupwise Chamfer loss as:
\begin{equation} 
\begin{split}L_{\text{Group-Chamfer}}(T(\mathcal{S}))&= \sum_{{S_i} \in \mathcal{S}} \sum_{{S_j} \neq {S_i}} L_{\text{Chamfer}}(T_i({S_i}),T_j({S_j})) \\
&+ \lambda \sum_{S_i \in \mathcal{S}}\sum_{x_i \in S_i}|dx_i|\\
\end{split}
\label{eq55}
\end{equation}
, where $\lambda$  is a balance term between the alignment loss and the deformation level. This balance term can be chosen as a fixed hyper-parameter or can be dynamically chosen during the optimization process. 

\subsection{Optimization Strategy}\label{sc_optim}
Algorithm 1 gives an illustration of the proposed method for groupwise point set registration. We notice that the step 3 to 8 in Algorithm 1 can be computed using GPUs. Learning rate decays from 0.001 to 0.0001 for the decoder network and GLDs during the first 100 steps. Adam optimizer is utilized for model optimization. We need to balance the alignment loss with the geometric distance of the drifts. Otherwise, one trivial solution can be easily reached to align all shapes towards one dot for example. The model is tested on a single Tesla P100 GPU. Our model usually requires 100 to 500 optimization steps until convergence. In the paper, we choose latent vector $Z_i$ with a dimension of 256 for each group ${\mathcal{S}_i}$. For the decoder network described in equation 3, we use 3 MLP layers with dimension (258,128,64) to regress the input $({S}, Z)$ into a 2-d/3-d drifts flow field. 

\section{Experiments}
In this experimental section, we demonstrate the performance of our model for groupwise point set registration. In section \ref{part1}, we demonstrate the performance of our model on aligning a single group of 2D synthesized point sets and 3D real-world point sets and we compare our model with state-of-the-art methods. In section \ref{part2}, we test our model on groupwise registration of a large number of groups of 3D real-world point sets. 

\subsection{Registration of single group of shapes}\label{part1}
In this section, we demonstrate the performance of our model for aligning one single group of 2D/3D shapes. In part \ref{part11}, we conduct extensive experiments using a group of 2D synthesized datasets. In part \ref{part12}, we demonstrate the performance of our model for aligning a group of 3D real-world point sets. 

\subsubsection{Experiments on 2D synthesized dataset} \label{part11}
In this section, we firstly demonstrate our model's groupwise point set registration performance on 2D synthesized dataset and compare our model with previous state-of-the-art methods \cite{giraldo2017group}. We conduct five experiments in this section to consecutively evaluate our model's performance on both unbiased and biased 2D data.\\

\begin{table*}
\begin{center}
\begin{tabular}{cccccc}
\hline
Def. level&Holder&IP&Norm-IP&Ours \\
\hline
0.2 &0.0106&0.0071&0.007&\textbf{0.00017} \\
0.4 &0.0426&0.0536&0.0523&\textbf{0.0019}\\
0.6 &0.1403&0.0929&0.0893&\textbf{0.0032} \\
\hline
\end{tabular}
\end{center}
\caption{Quantitative testing performance on fish-shaped point sets at different deformation levels. A lower number indicates better groupwise registration performance.}
\label{t1}
\end{table*}
\begin{table*}
\begin{center}
\begin{tabular}{ccccc}
\hline
Holder&IP&Norm-IP&Ours \\
\hline
88s&85s&78s&23s \\
\hline
\end{tabular}
\end{center}
\caption{Time for completing 1000 steps for groupwise registration of seven fish-shaped point sets at deformation level 0.4.}
\label{t2}
\end{table*}

\noindent \textbf{Dataset. } For synthesizing 2D deformable point sets, we simulate non-rigid geometric transformation on the normalized raw point sets by Thin Plate Spline (TPS) \cite{bookstein1989principal} transformation with a given deformation level. The deformation level is defined as the perturbing degree of controlling points in TPS. Specifically, given the deformation level at $l$, a Gaussian random shift with zero-mean and $2l$ standard deviation is generated to perturb the controlling points. When the deformation level is as low as in \cite{giraldo2017group}, previous methods can achieve nearly perfect alignment results. For better comparison, we increase the deformation level and therefore bring challenges for this task in comparison to original settings in \cite{giraldo2017group}.

\begin{figure}
\begin{center}
\includegraphics[width=8.5cm]{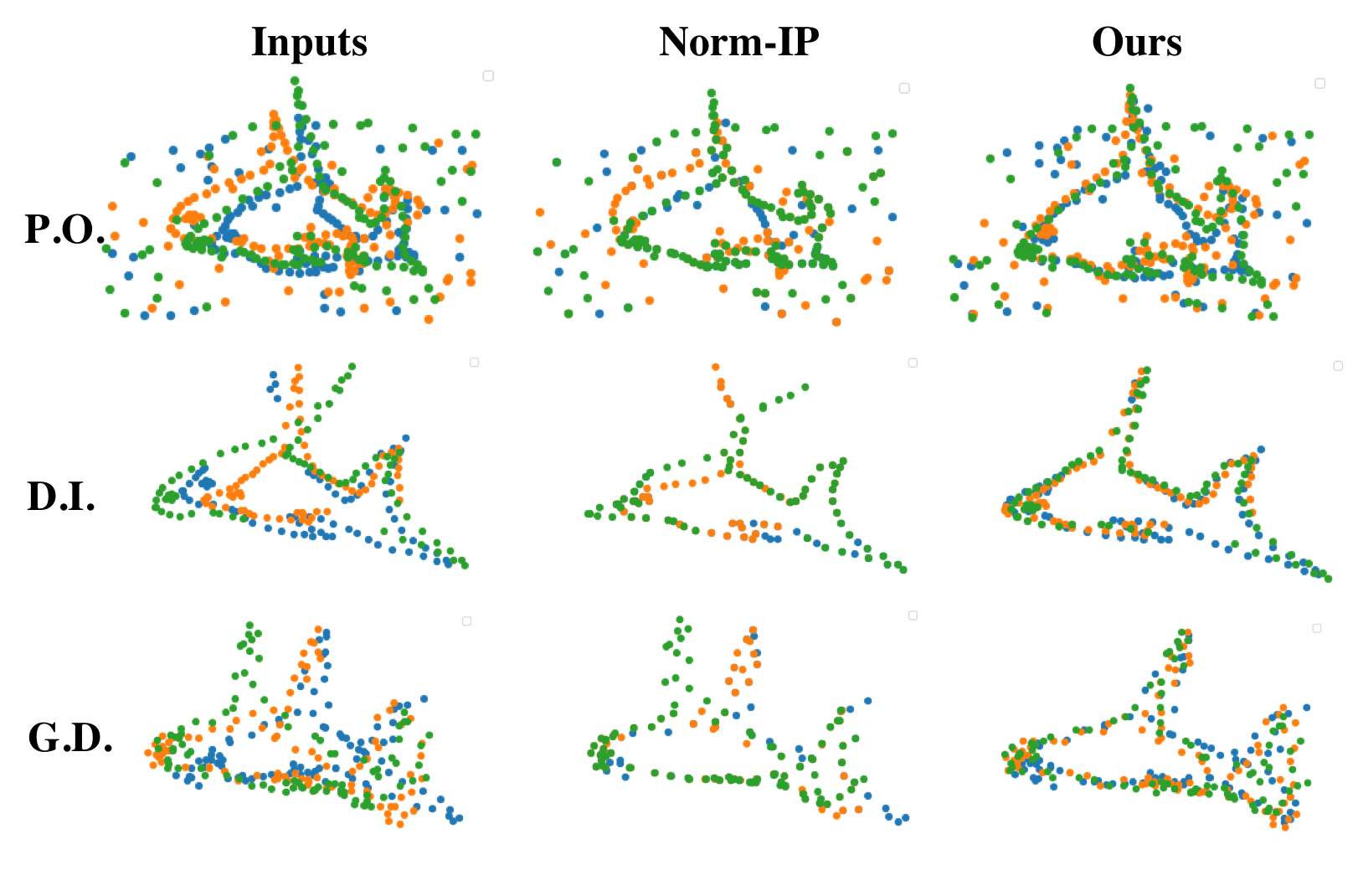}
\end{center}
\caption{The qualitative registration results of aligning fish-shaped point sets at different deformation levels. Inputs include seven point sets in different colors in presence of various non-rigid deformation.}
\label{exp2}
\end{figure}

For synthesizing 2D point sets, three different types of noise are added to the input 2D point set. We call these three types of noise Point Outlier (P.O.) noise, Data Incompleteness (D.I.) noise, and Gaussian Displacement (G.D.) noise. To prepare the Point Outlier (P.O.) noise for the shape (as shown in the first row of Figure \ref{exp2}), we simulate the outliers on the deformed point set by adding a certain number of Gaussian outliers. The P.O. noise level is defined as a ratio of Gaussian outliers and the entire target point set. To prepare the Data Incompleteness (D.I.) noise (as shown in the second row of Figure \ref{exp2}), we remove a certain number of neighboring points from the target point set. The D.I. noise level is defined as the percentage of the points removed from the entire target point set. To prepare the Gaussian Displacement (G.D.) noise dataset (as shown in the third row of Figure \ref{exp2}), we simulate the random displacement superimposed on a deformed point set by applying an increasing intensity of zero-mean Gaussian noise. The G.D. noise level is the standard deviation of Gaussian. For experiment 1, we randomly choose 7 synthesized point sets at deformation level 0.4 for demonstration of the intermediate alignment result during the optimization process. For experiment 2, we randomly choose 7 synthesized point sets at deformation from 0.2 to 0.6 as the dataset to evaluate models' groupwise registration performance on the unbiased dataset. For experiment 3, we randomly select 3 point sets at deformation level 0.4 and we separately add P.O noise with level 0.4, D.I. noise with level 0.2, and G.D. noise with level 0.05 to the selected shapes to further evaluate models' groupwise registration performance on the biased dataset. \\

\noindent \textbf{Settings. }As explained in section \ref{sc_optim}, we choose latent vector $Z$ with a dimension of 256. For the decoder, we use 3 MLP layers with dimension (258,128,64) to regress a 2-dimensional drifts flow field. 

\begin{figure}
\begin{center}
\includegraphics[width=8cm]{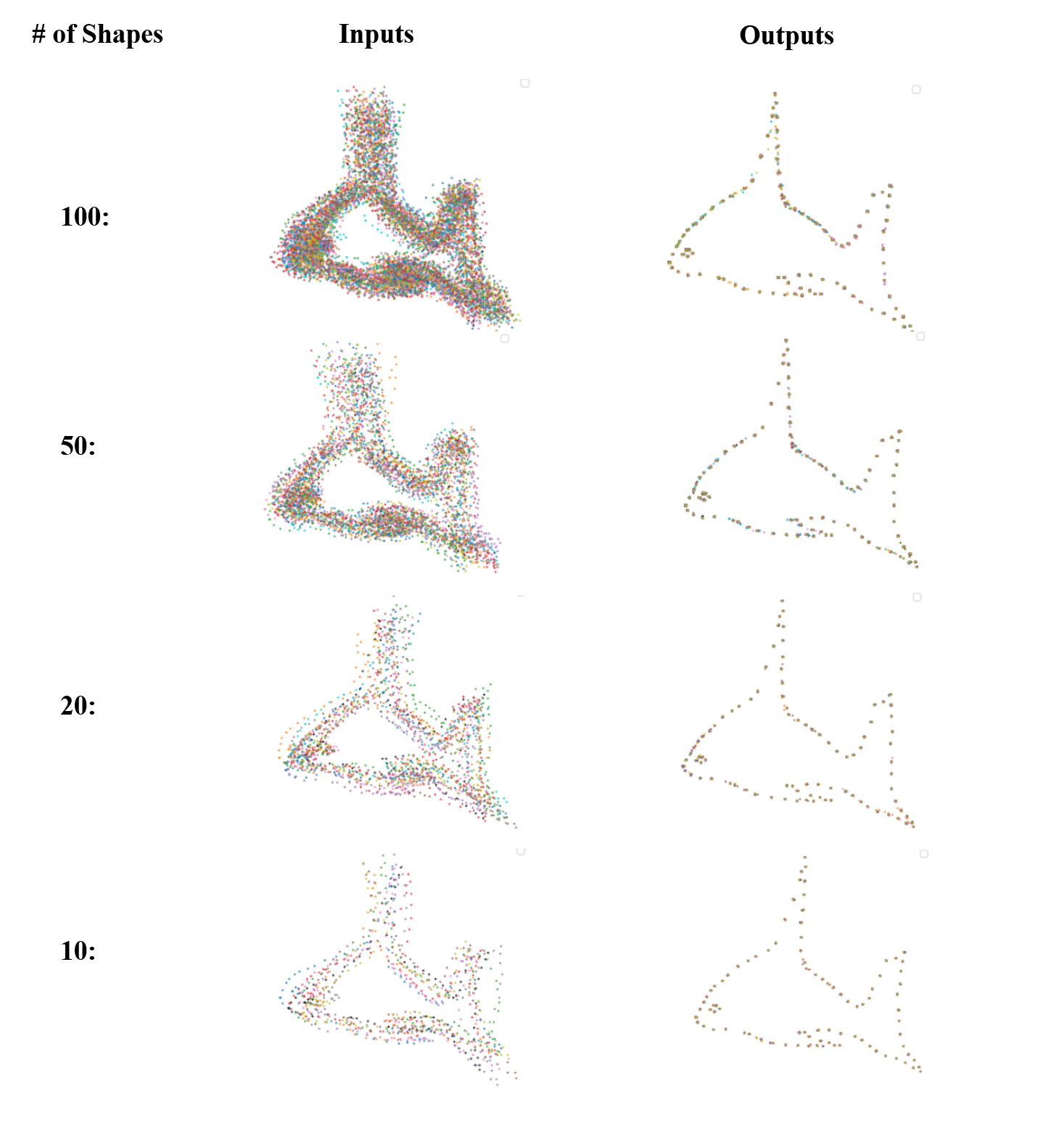}
\end{center}
\caption{The qualitative registration results of aligning different number of fish-shaped point sets. Inputs include 10 to 100 point sets in different colors in presence of various non-rigid deformation.}
\label{exp1111}
\end{figure}

\begin{itemize}
\item For Test 1, we examine the alignment process for a group of input point sets by the optimization steps. The inputs include 7 non-rigid deformed shapes and the deformation level of inputs is 0.4. For this test, we demonstrate the qualitative result in Figure \ref{steps}.\\

\item For Test 2, we examine the impact of initialization status on the final alignment result. For this test, the inputs include 7 non-rigid deformed fish shapes and the deformation level of inputs is set to 0.4. We rerun our experiments 50 times with different initialization and report the quantitative results.\\

\item For Test 3, we demonstrate the performance of our model for registering a large number of non-rigid deformed shapes at the same time. For this test, we set the deformation level at 0.2. We randomly generalize a set of 10, 20, 50, 100 shapes as inputs and we allow a maximum of 500 steps for each optimization process. The quantitative result is demonstrated in Table \ref{t1111} and the qualitative result is presented in Figure \ref{exp1111}.\\

\item For Test 4, we compare the performance of our model for groupwise registration with the state-of-the-art models on a group of synthesized shapes with different deformation levels. We consecutively increase the deformation level from 0.2 to 0.6. We compare our model with the state-of-the-art models: Holder, IP, and Norm-IP \cite{giraldo2017group}. Norm-IP is the previous state-of-the-art model. For quantitative results, we use normalized outputs' groupwise Chamfer Distance (C.D.) in formula 5 as our metric, which is shown in Table \ref{t1}. We demonstrate the qualitative result in Figure \ref{exp1}. Furthermore, for comparison of algorithm efficiency between our model with previous methods, we provide the running time for reference in Table \ref{t2}.\\

\item For Test 5, we demonstrate the performance of our model on biased data. The dataset is prepared as explained in the previous section. For this test, we use the previous state-of-the-art model Norm-IP as our baseline model for comparison. The qualitative results are demonstrated in Figure \ref{exp2}.
\end{itemize}

\noindent \textbf{Results of Test 1. }As shown in Figure \ref{steps}, we notice that after 100 steps, the main parts of the input shapes can be well aligned. During the experiment, we notice that the convergence speed after 100 steps becomes comparatively slow, which can be observed in Figure \ref{steps} from as well. All shapes are aligned perfectly after reaching 300 steps (convergence). We need to point out that the middle shape (template) is different from any input shape and is automatically determined during the optimization process.\\

\noindent \textbf{Results of Test 2. }The initialization problem is important for groupwise alignment. For aligning the same group of 7 fish shapes, our model achieved the average normalized groupwise Chamfer distance 0.0018 for 50 different initialization status. The standard deviation of the Chamfer distance for 50 observations is 0.00016. The highest score of these observations is 0.0022 and the lowest score of these observations is 0.0014. Therefore, we notice that the initialization status has a small impact on our model's final alignment performance.\\

\noindent \textbf{Results of Test 3. } As shown in Table \ref{t1111}, the quantitative results show that our model achieved similar groupwise Chamfer distance for aligning 10 to 100 shapes. The groupwise Chamfer distance achieved for aligning 100 shapes is even the best among these cases. Also, the computation time is longer for aligning 100 shapes (5m11s) than 10 shapes (38s) as expected. In Figure \ref{exp1111}, we further notice that all the shapes are perfectly aligned for 10 to 100 input shapes without significant differences.\\

\begin{table}[h]
\begin{center}
\begin{tabular}{cccc}
\hline
Number of Shapes&Groupwise C.D.&Time \\
\hline
10 &0.00022&38s \\
20&0.00026&1m06s \\
50&0.00031&2m33s \\
100&0.00011&5m11s \\
\hline
\end{tabular}
\end{center}
\caption{Quantitative testing performance and time for running 500 steps for aligning different number of fish-shaped point sets.}
\label{t1111}
\end{table}

\noindent \textbf{Results of Test 4. }As shown in Figure \ref{exp1}, the inputs of point sets are shown in the first column. The outputs of our model and comparing models are consecutively demonstrated from column two to five. When the input data at the lower deformation level, Norm-IP demonstrates good performance for most shapes except the blue one. When deformation level increases, by leveraging the power of neural networks, our model outperforms the comparing models. From the quantitative results shown in Table \ref{t1}, our model demonstrates significantly better performance than all comparing methods under different deformation levels. The alignment C.D. distance of our model is a magnitude better than the current state-of-the-art Norm-IP model. Moreover, regarding the algorithm efficiency, as shown in Table \ref{t2}, the state-of-the-art Norm-IP models require 78s to register a group of 7 fish shapes, while our model only takes 23s to accomplish the registration process. When dealing with real-world 3D point sets, efficiency improvement becomes even more significant, which will be shown in the next section.\\

\noindent \textbf{Results of Test 5. }In this experiment, we compare our performance of our model with the previous state-of-the-art model Norm-IP on aligning a group of point sets in the presence of various noise patterns. As shown in Figure \ref{exp2}, the input point sets are shown in the first column. The outputs of model Norm-IP are demonstrated in the middle column and our results are shown in the last column. Under P.O. noise, we can see that the orange color fish is mistakenly aligned with the other two fishes for Norm-IP, but our model can generally align them. Under D.I., our model can almost perfectly align the input group of shapes but the Norm-IP model converges into two patterns. Under G.D., in comparison with Norm-IP, our model successfully aligned the top part of fishes. In all cases, we notice that our model achieves better results. 

\subsubsection{Experiments on 3D dataset} \label{part12}
In this section, we demonstrate the effectiveness of our model for groupwise point set registration on single group of 3D real-world shapes.\\

\noindent \textbf{Dataset. }We conduct our experiment on the chair category of ShapeNet \cite{chang2015shapenet} dataset. We randomly select 3 shapes from this category as the inputs as shown in Figure \ref{exp3}. Each shape contains 2048 uniformly sampled points. \\

\noindent \textbf{Setting. }As explained in section \ref{sc_optim}, we choose latent vector $Z$ with a dimension of 256 for each group. For the decoder network, we use a 3-layer MLP with dimensions of (258,128,64) to generate the 3D drifts. 

\begin{itemize}
\item For Test 1, we compare our model with Norm-IP \cite{giraldo2017group} which is the previous state-of-the-art model. Similarly to the previous section, we use normalized groupwise Chamfer Distance (D.C.) as the evaluation. The results are listed in the first row of Table \ref{t3}. We report the running time to demonstrate our model's efficiency in the second row of Table \ref{t3}. Qualitative results are shown in Figure \ref{exp3}.\\
\begin{figure}[h]
\begin{center}
\includegraphics[width=8cm]{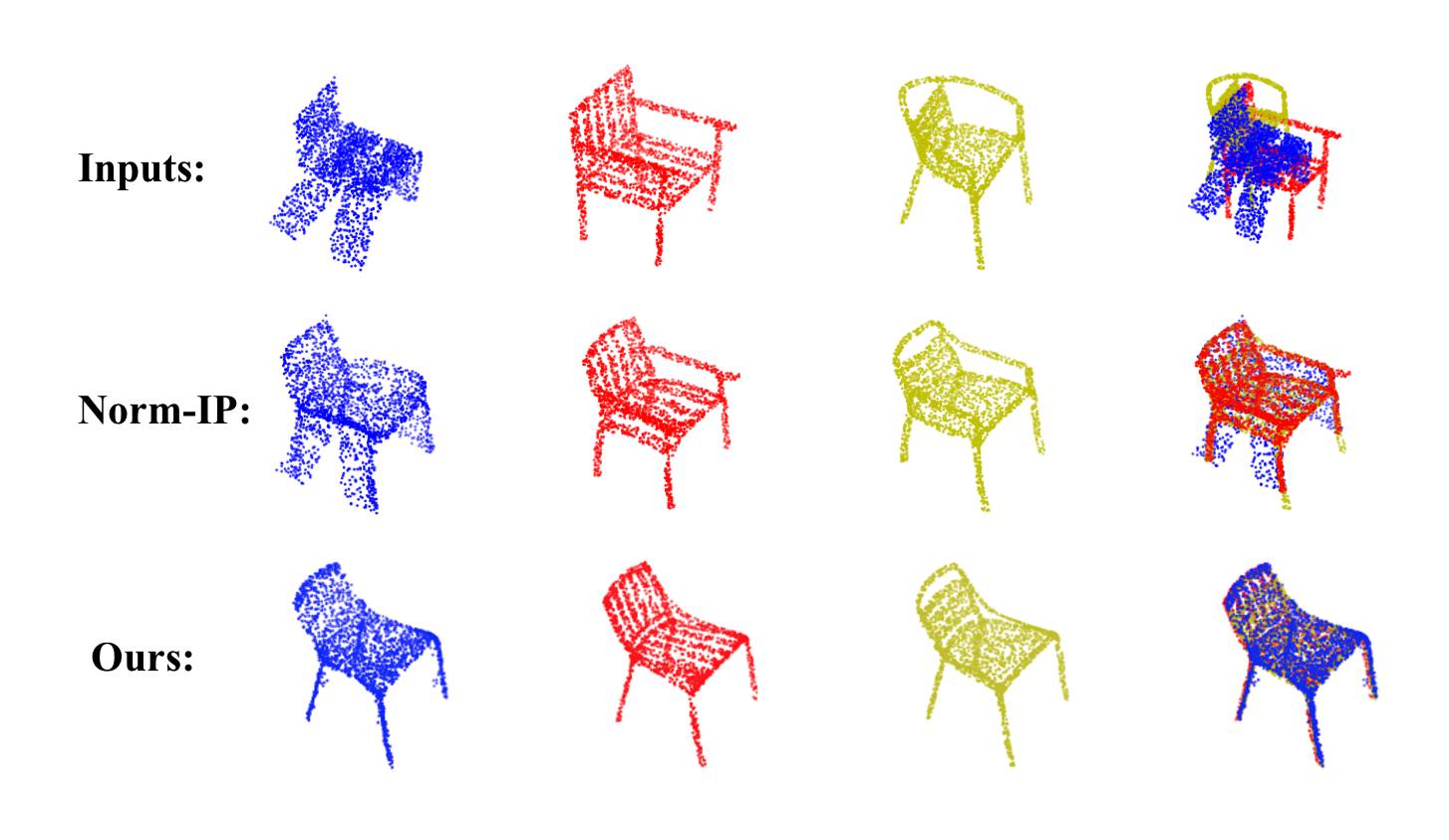}
\end{center}
\caption{The qualitative registration results for 3D real-world point sets. Inputs include three shapes (in different
colors) from chair category of ShapeNet \cite{chang2015shapenet} dataset.}
\label{exp3}
\end{figure}

\begin{table}[h]
\begin{center}
\begin{tabular}{cccc}
\hline
Methods&Norm-IP&Ours \\
\hline
groupwise C.D. &0.0421&\textbf{0.0040} \\
Time (100 steps)&4328s&\textbf{16s} \\
\hline
\end{tabular}

\end{center}
\caption{Quantitative testing performance and time for running 100 steps for 3D point set registration. For achieving the results with reported groupwise C.D. in this table, we run Norm-IP for 100 steps and GP-Aligner for 200 steps.}
\label{t3}
\end{table}

\begin{figure}[h]
\begin{center}
\includegraphics[width=8cm]{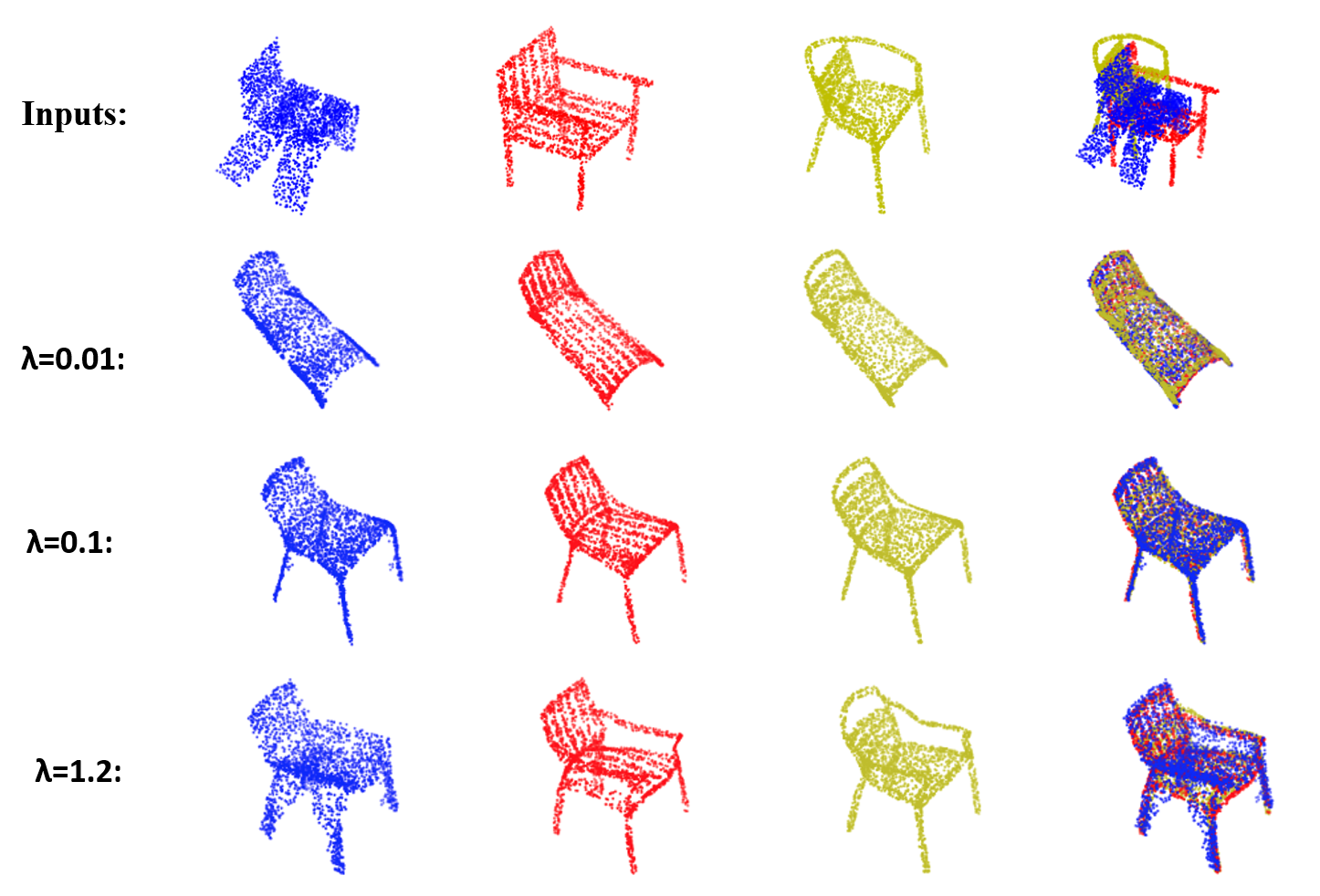}
\end{center}
\caption{The qualitative registration results for 3D real-world point sets with different regularization weights. Inputs include three shapes (in different colors) from chair category of ShapeNet \cite{chang2015shapenet} dataset.}
\label{exp3333}
\end{figure}

\begin{table}[h]
\begin{center}
\begin{tabular}{cc}
\hline
Reg.Term& Groupwise D.C.\\
\hline
0.01&0.0013\\
0.1&0.0040\\
1.2&0.0236\\
\hline
\end{tabular}
\end{center}
\caption{Quantitative performance for 3D point set registration with different regularization weights.}
\label{t3333}
\end{table}
\item For Test 2, we analyze the impacts of the scale of balance term for regularization as explained in \ref{sc_loss} on the final groupwise registration performance. For this experiment, we use the same setting as in Test 1 but we set the balance term $\lambda$ to 0.01, 0.1, and 1.2. We report the alignment performance in Table \ref{t3333} and the qualitative results are shown in Figure \ref{exp3333}.\\
\end{itemize}

\begin{figure*}
\begin{center}
\includegraphics[width=18cm]{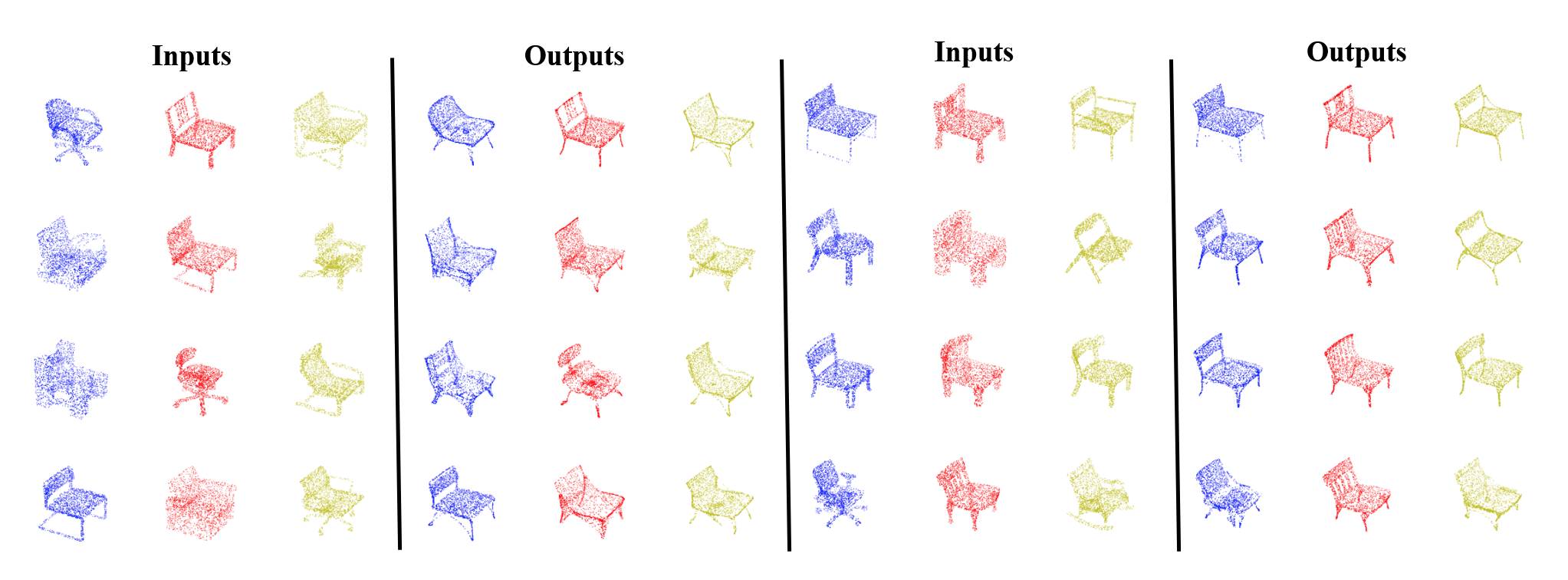}
\end{center}
\caption{The qualitative registration results for registration of multiple 3D real-world point sets in Chair Category of ShapeNet \cite{chang2015shapenet} dataset. We randomly select one entire batch of 8 groups for demonstration.}
\label{exp2233}
\end{figure*}
\noindent \textbf{Results of Test 1. }As shown in Figure \ref{exp3}, our GP-Aligner successfully transformed three input shapes to one common middle shape (last column). We can notice that the transformed shapes belong to one style and the deformation is reasonable. The results of Norm-IP also demonstrate a reasonable transformation as shown in the second row of Figure \ref{exp3}. Our model achieves a magnitude better performance than the Norm-IP model as indicated by the groupwise D.C. in Table \ref{t3}. More importantly, as shown in the second row of Table \ref{t3}, Norm-IP requires 4328 seconds to run 100 steps, while our model only needs 16 seconds. This difference in model efficiency can be even larger when increasing the number of shapes or the number of sampled points in one shape. \\

\noindent \textbf{Results of Test 2. }In Figure \ref{exp3333}, we show the registration results of our model under different settings of the regularization weight. The first three columns show the aligned result of each shape and the last column shows the overlapped shapes. As shown in Figure \ref{exp3333}. For a lower regularization value such as 0.01, the topological structure of shapes can be dramatically deformed and lose their original semantic meanings. For example, all the legs of the chairs are missing after registration. But the deformation field is still coherent without local disturbances based on the difference between inputs and outputs. With a large balance weight such as 1.2, the deformation freedom is limited and we can see the resulting groupwise Chamfer distance is much higher from Table \ref{t3333} in comparison to the other two cases. Therefore, there is always a trade-off between the alignment loss and the regularization level. Higher regularization level indicates a better topological consistency between input and output shapes but worse alignment loss among transformed shapes. \\

\begin{figure*}
\begin{center}
\includegraphics[width=13cm]{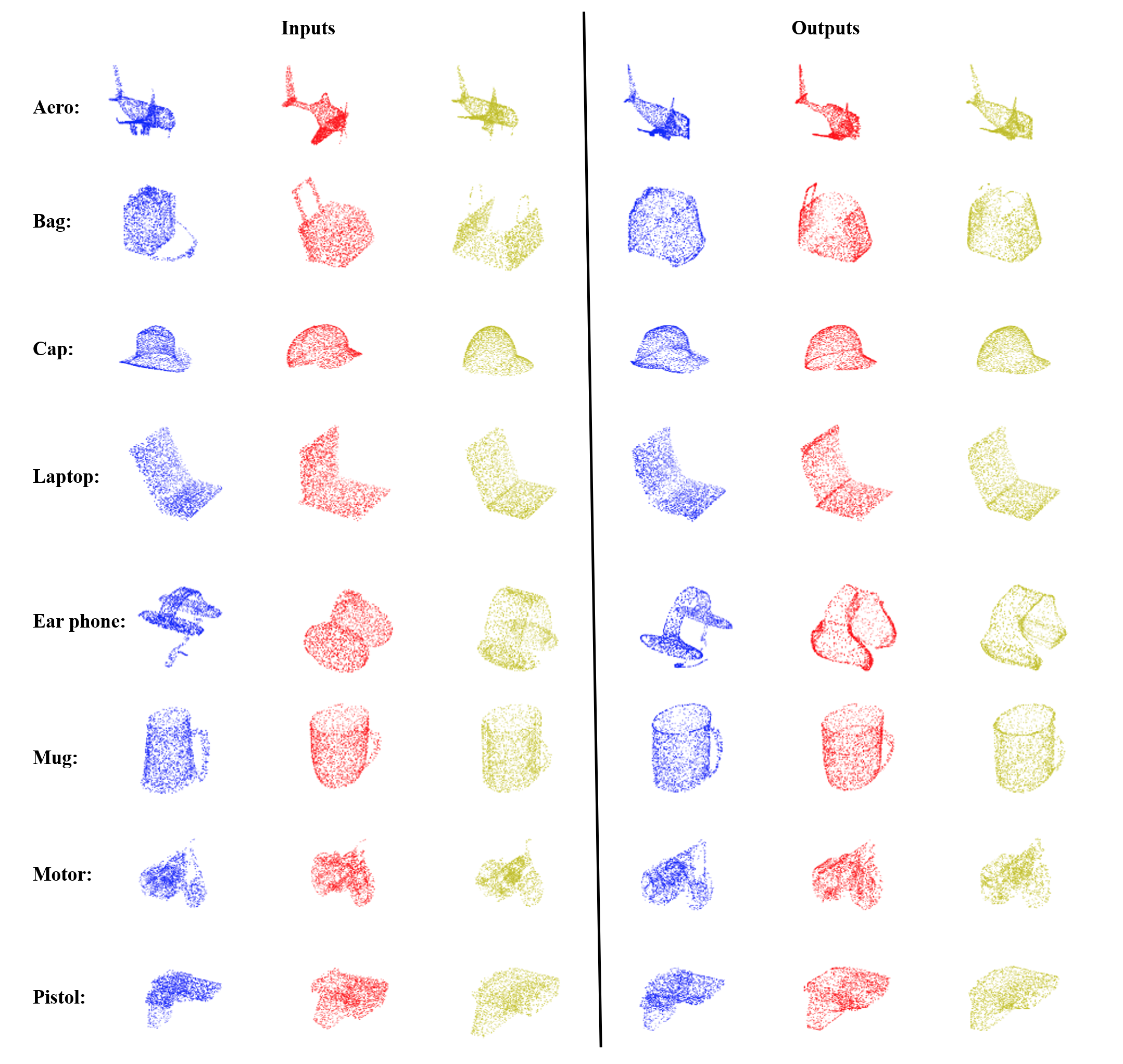}
\end{center}
\caption{The qualitative registration results for registration of multiple groups of 3D real-world point sets in selected Categories of ShapeNet \cite{chang2015shapenet} dataset. We randomly select one case from 100 groups for demonstration for each category}
\label{exp2234}
\end{figure*}
\begin{figure*}
\begin{center}
\includegraphics[width=9cm]{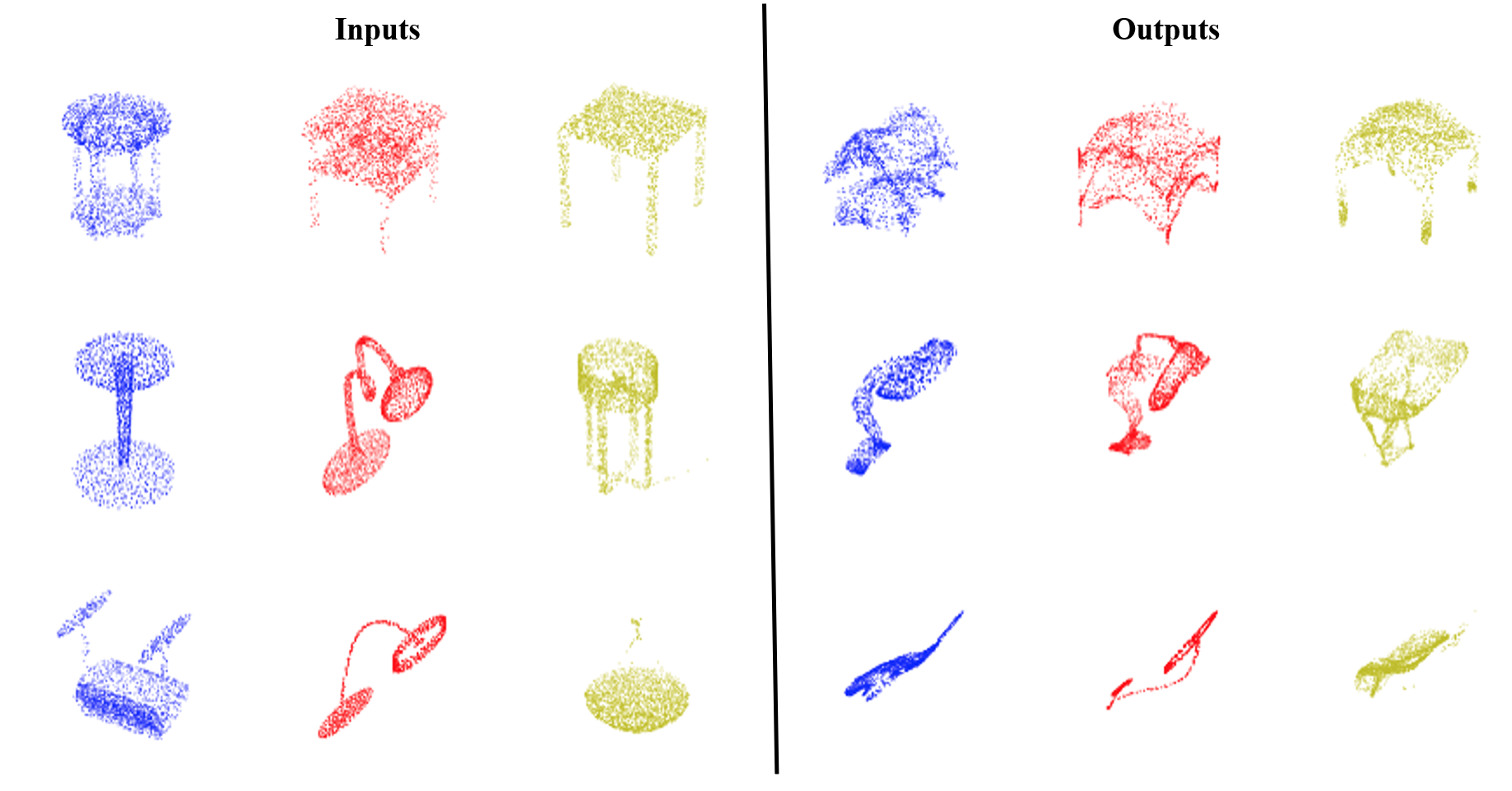}
\end{center}
\caption{The selected failed cases for registration of multiple groups of 3D real-world point sets.}
\label{exp2235}
\end{figure*}

\begin{table}[h]
\begin{center}
\begin{tabular}{ccc}
\hline
Categories&Groupwise D.C.&Groupwise D.C.\\
&Before Alignment& After Alignment\\
\hline
aero&0.1719&0.0232\\
bag&0.3249&0.0173\\
cap&0.1971&0.0138\\
car&0.0638&0.0313\\
chair&0.2432&0.0249\\
ear phone&0.6240&0.0233\\
guitar&0.1135&0.0194\\
knife&0.1663&0.0177\\
lamp&1.5620&0.0479\\
laptop&0.0548&0.0158\\
motor&0.1502&0.0405\\
mug&0.1301&0.0194\\
pistol&0.1105&0.0268\\
rocket&0.4457&0.0151\\
skate&0.1409&0.0149\\
table&0.4844&0.0188\\
\hline
Average &0.3115&0.0231\\
Running Time &27min&-\\
\hline
\end{tabular}
\end{center}
\caption{Quantitative performance and time for aligning 100 groups of 3D real-world point sets from all the categories of ShapeNet dataset. We report the groupwise Chamfer distance for cases before and after alignment.}
\label{t2233}
\end{table}
\noindent \textbf{Discussion. }The deformation of a group of 3D shapes towards a middle shape has a wide range of applications in practice. To align a group of real-world 3D point sets as shown in this experiment, our method shows significantly better performance in comparison to the state-of-the-art Norm-IP method. By leveraging the power of GPU computation and an efficient optimization algorithm, our method also shows significantly better computation efficiency. Moreover, unlike previous learning-based methods, the proposed network can be regarded as an optimization algorithm, which does not require any labeled training dataset. The entire alignment process is completed during a one-stage optimization process. These characteristics make our method more applicable to solving real-world industrial problems. The reasonable deformation fields link a group of 3D shapes together with reasonable correspondences. The groupwise registration result for real-world 3D shapes can be further used for tasks such as groupwise segmentation, correspondence, and so on. Moreover, our method can be easily extended to the registration of a group of volumetric 3D shapes by replacing the decoder with 3D deconvolutional layers. Thus, our model can be used for groupwise medical images registration. Future work will extend this method for registration of a group of volumetric shapes in the domain of medical imaging. In the next part, we extend our model to register a large number of groups at the same time. For aligning multiple groups, our system with a complex deep learning-based feature exaction module with the newly designed optimizable group-based latent vector demonstrates dramatic improvement in efficiency. 

\subsection{Registration of multiple groups}\label{part2}
In this section, we demonstrate the performance of our model for the registration of multiple groups of 3D real-world datasets.\\

\noindent \textbf{Dataset. }We conduct experiments on 14 categories of ShapeNet \cite{chang2015shapenet} dataset. For preparing the multiple groups dataset, we randomly select 100 groups of shapes where each group includes 3 randomly selected shapes for each shape category. Each shape contains 2048 uniformly sampled points.\\

\noindent \textbf{Setting. }As explained in section \ref{sc_optim}, we choose latent vector $Z_i$ with a dimension of 256 for each group. For the decoder network, we use a 3-layer MLP with dimensions of (258,128,64) to generate the 3D drifts. Due to a large number of groups, the previous methods do not apply to this experiment. We use normalized groupwise Chamfer Distance (D.C.) as the evaluation method and we list the quantitative results for the inputs before alignment and outputs after alignment in Table \ref{t2233}. We report the running time for every 100 groups to demonstrate the efficiency in Table \ref{t2233}. Qualitative results are shown in Figure \ref{exp2233} for one randomly selected batch of groups of chair category. The results of the other categories are shown in Figure \ref{exp2234}. In Figure \ref{exp2235}, we selected a few failed cases to show that our model may fail to maintain semantic meaning after deformations, especially when the topological structure dramatically varies among the shapes in the group.\\

\noindent \textbf{Results. }In Figure \ref{exp2233}, we show the registration results of 8 groups of input shapes. From the results, one can see that our model produces satisfying alignments for each group of shapes, even though the topological structures are quite different for some groups such as the first case in row 3. The legs in this group are well-aligned from three shapes with large structural variations. As shown in Figure \ref{exp2234}, for most other categories, our GP-Aligner can successfully align the three input shapes to one common middle shape. We also notice that for categories with high similarity in topological structure among input shapes, e.g., the laptop category, the alignment performance is slightly better as shown in Table \ref{t2233}. For categories with huge topological variations, e.g., the lamp category, our model achieves worse final alignment results than other categories. Figure \ref{exp2235} shows a few failed cases of our model. In this figure, for the lamp case in row 2, the third input lamp has a different structure from the first two lamps. Therefore, the deformed shape looks weird. For the third example, all three input lamps are completely different, we cannot expect reasonable alignment for this case.  

As regards the registration efficiency, our model can accomplish the registration tasks of 100 groups of 3D point sets in 27 minutes. We do not compare the efficiency with the comparing models due to the quite long computation time required by these methods.

\section{Conclusion}
In this paper, we introduce a novel method, called GP-Aligner, for non-rigid groupwise point set registration. The proposed method is built upon a deep neural network architecture and can perform the registration for groups of point sets in an unsupervised way. To avoid the explicit design of a shape feature encoding network, our model leverages a model-free learnable latent descriptor to characterize the group relationship. An optimizable Group Latent Descriptor (GLD) is designed to characterize the groupwise relationship among a group of point sets. The GLD is randomly initialized from a Gaussian distribution and optimized along with the network parameters during the optimization process. We conduct extensive experiments for both synthesized shapes and real-world 3D shapes. Experimental results show that our GP-Aligner model can achieve superior registration performance as well as running efficiency. We also demonstrate the robustness of our model under different types of input noises. For registering a large number of groups of shapes, our model shows satisfying registration performance and remarkable running efficiency.

\ifCLASSOPTIONcaptionsoff
  \newpage
\fi

\bibliographystyle{IEEEtran}
\bibliography{egbib}

\begin{thebibliography}{10}
\providecommand{\url}[1]{#1}
\csname url@samestyle\endcsname
\providecommand{\newblock}{\relax}
\providecommand{\bibinfo}[2]{#2}
\providecommand{\BIBentrySTDinterwordspacing}{\spaceskip=0pt\relax}
\providecommand{\BIBentryALTinterwordstretchfactor}{4}
\providecommand{\BIBentryALTinterwordspacing}{\spaceskip=\fontdimen2\font plus
\BIBentryALTinterwordstretchfactor\fontdimen3\font minus
  \fontdimen4\font\relax}
\providecommand{\BIBforeignlanguage}[2]{{%
\expandafter\ifx\csname l@#1\endcsname\relax
\typeout{** WARNING: IEEEtran.bst: No hyphenation pattern has been}%
\typeout{** loaded for the language `#1'. Using the pattern for}%
\typeout{** the default language instead.}%
\else
\language=\csname l@#1\endcsname
\fi
#2}}
\providecommand{\BIBdecl}{\relax}
\BIBdecl

\bibitem{myronenko2009image}
A.~Myronenko and X.~Song, ``Image registration by minimization of residual
  complexity,'' 2009.

\bibitem{ma2016non}
J.~Ma, J.~Zhao, and A.~L. Yuille, ``Non-rigid point set registration by
  preserving global and local structures,'' \emph{IEEE Transactions on image
  Processing}, vol.~25, no.~1, pp. 53--64, 2016.

\bibitem{maintz1998survey}
J.~A. Maintz and M.~A. Viergever, ``A survey of medical image registration,''
  \emph{Medical image analysis}, vol.~2, no.~1, pp. 1--36, 1998.

\bibitem{wang2019deep}
Y.~Wang and J.~M. Solomon, ``Deep closest point: Learning representations for
  point cloud registration,'' \emph{arXiv preprint arXiv:1905.03304}, 2019.

\bibitem{myronenko2007non}
A.~Myronenko, X.~Song, and M.~A. Carreira-Perpin{\'a}n, ``Non-rigid point set
  registration: Coherent point drift,'' in \emph{Advances in Neural Information
  Processing Systems}, 2007, pp. 1009--1016.

\bibitem{ma2014robust}
J.~Ma, J.~Zhao, J.~Tian, A.~L. Yuille, and Z.~Tu, ``Robust point matching via
  vector field consensus.'' \emph{IEEE Trans. image processing}, vol.~23,
  no.~4, pp. 1706--1721, 2014.

\bibitem{ling2005deformation}
H.~Ling and D.~W. Jacobs, ``Deformation invariant image matching,'' in
  \emph{Computer Vision, 2005. ICCV 2005. Tenth IEEE International Conference
  on}, vol.~2.\hskip 1em plus 0.5em minus 0.4em\relax IEEE, 2005, pp.
  1466--1473.

\bibitem{viergever2016survey}
M.~A. Viergever, J.~A. Maintz, S.~Klein, K.~Murphy, M.~Staring, and J.~P.
  Pluim, ``A survey of medical image registration--under review,'' 2016.

\bibitem{besl1992method}
P.~J. Besl and N.~D. McKay, ``Method for registration of 3-d shapes,'' in
  \emph{Sensor Fusion IV: Control Paradigms and Data Structures}, vol.
  1611.\hskip 1em plus 0.5em minus 0.4em\relax International Society for Optics
  and Photonics, 1992, pp. 586--607.

\bibitem{sotiras2013deformable}
A.~Sotiras, C.~Davatzikos, and N.~Paragios, ``Deformable medical image
  registration: A survey,'' \emph{IEEE transactions on medical imaging},
  vol.~32, no.~7, pp. 1153--1190, 2013.

\bibitem{liu2019flownet3d}
X.~Liu, C.~R. Qi, and L.~J. Guibas, ``Flownet3d: Learning scene flow in 3d
  point clouds,'' in \emph{Proceedings of the IEEE Conference on Computer
  Vision and Pattern Recognition}, 2019, pp. 529--537.

\bibitem{balakrishnan2018unsupervised}
G.~Balakrishnan, A.~Zhao, M.~R. Sabuncu, J.~Guttag, and A.~V. Dalca, ``An
  unsupervised learning model for deformable medical image registration,'' in
  \emph{Proceedings of the IEEE Conference on Computer Vision and Pattern
  Recognition}, 2018, pp. 9252--9260.

\bibitem{rocco2017convolutional}
I.~Rocco, R.~Arandjelovic, and J.~Sivic, ``Convolutional neural network
  architecture for geometric matching,'' in \emph{Proc. CVPR}, vol.~2, 2017.

\bibitem{chen2019arbicon}
J.~Chen, L.~Wang, X.~Li, and Y.~Fang, ``Arbicon-net: Arbitrary continuous
  geometric transformation networks for image registration,'' in \emph{Advances
  in Neural Information Processing Systems}, 2019, pp. 3410--3420.

\bibitem{che2019deep}
T.~Che, Y.~Zheng, J.~Cong, Y.~Jiang, Y.~Niu, W.~Jiao, B.~Zhao, and Y.~Ding,
  ``Deep group-wise registration for multi-spectral images from fundus
  images,'' \emph{IEEE Access}, vol.~7, pp. 27\,650--27\,661, 2019.

\bibitem{wang2020unsupervised}
L.~Wang, Y.~Shi, X.~Li, and Y.~Fang, ``Unsupervised learning of global
  registration of temporal sequence of point clouds,'' \emph{arXiv preprint
  arXiv:2006.12378}, 2020.

\bibitem{Ding_2019_CVPR}
L.~Ding and C.~Feng, ``Deepmapping: Unsupervised map estimation from multiple
  point clouds,'' in \emph{The IEEE Conference on Computer Vision and Pattern
  Recognition (CVPR)}, June 2019.

\bibitem{giraldo2017group}
L.~G.~S. Giraldo, E.~Hasanbelliu, M.~Rao, and J.~C. Principe, ``Group-wise
  point-set registration based on renyi's second order entropy,'' in \emph{2017
  IEEE Conference on Computer Vision and Pattern Recognition (CVPR)}.\hskip 1em
  plus 0.5em minus 0.4em\relax IEEE, 2017, pp. 2454--2462.

\bibitem{chen2010group}
T.~Chen, B.~C. Vemuri, A.~Rangarajan, and S.~J. Eisenschenk, ``Group-wise
  point-set registration using a novel cdf-based havrda-charv{\'a}t
  divergence,'' \emph{International journal of computer vision}, vol.~86,
  no.~1, p. 111, 2010.

\bibitem{wang2006groupwise}
F.~Wang, B.~C. Vemuri, and A.~Rangarajan, ``Groupwise point pattern
  registration using a novel cdf-based jensen-shannon divergence,'' in
  \emph{2006 IEEE Computer Society Conference on Computer Vision and Pattern
  Recognition (CVPR'06)}, vol.~1.\hskip 1em plus 0.5em minus 0.4em\relax IEEE,
  2006, pp. 1283--1288.

\bibitem{chui2003new}
H.~Chui and A.~Rangarajan, ``A new point matching algorithm for non-rigid
  registration,'' \emph{Computer Vision and Image Understanding}, vol.~89, no.
  2-3, pp. 114--141, 2003.

\bibitem{chui2004unsupervised}
H.~Chui, A.~Rangarajan, J.~Zhang, and C.~M. Leonard, ``Unsupervised learning of
  an atlas from unlabeled point-sets,'' \emph{IEEE Transactions on Pattern
  Analysis and Machine Intelligence}, vol.~26, no.~2, pp. 160--172, 2004.

\bibitem{chui2000new}
H.~Chui and A.~Rangarajan, ``A new algorithm for non-rigid point matching,'' in
  \emph{Computer Vision and Pattern Recognition, 2000. Proceedings. IEEE
  Conference on}, vol.~2.\hskip 1em plus 0.5em minus 0.4em\relax IEEE, 2000,
  pp. 44--51.

\bibitem{wang2019non}
L.~Wang, J.~Chen, X.~Li, and Y.~Fang, ``Non-rigid point set registration
  networks,'' \emph{arXiv preprint arXiv:1904.01428}, 2019.

\bibitem{wang2019coherent}
L.~Wang, X.~Li, J.~Chen, and Y.~Fang, ``Coherent point drift networks:
  Unsupervised learning of non-rigid point set registration,'' \emph{arXiv
  preprint arXiv:1906.03039}, 2019.

\bibitem{rasoulian2012group}
A.~Rasoulian, R.~Rohling, and P.~Abolmaesumi, ``Group-wise registration of
  point sets for statistical shape models,'' \emph{IEEE transactions on medical
  imaging}, vol.~31, no.~11, pp. 2025--2034, 2012.

\bibitem{siebert2020deep}
H.~Siebert and M.~P. Heinrich, ``Deep groupwise registration of mri using
  deforming autoencoders,'' in \emph{Bildverarbeitung f{\"u}r die Medizin
  2020}.\hskip 1em plus 0.5em minus 0.4em\relax Springer, 2020, pp. 236--241.

\bibitem{qi2017pointnet}
C.~R. Qi, H.~Su, K.~Mo, and L.~J. Guibas, ``Pointnet: Deep learning on point
  sets for 3d classification and segmentation,'' \emph{Proc. Computer Vision
  and Pattern Recognition (CVPR), IEEE}, vol.~1, no.~2, p.~4, 2017.

\bibitem{yuille1989mathematical}
A.~L. Yuille and N.~M. Grzywacz, ``A mathematical analysis of the motion
  coherence theory,'' \emph{International Journal of Computer Vision}, vol.~3,
  no.~2, pp. 155--175, 1989.

\bibitem{fan2017point}
H.~Fan, H.~Su, and L.~J. Guibas, ``A point set generation network for 3d object
  reconstruction from a single image.'' in \emph{CVPR}, vol.~2, no.~4, 2017,
  p.~6.

\bibitem{bookstein1989principal}
F.~L. Bookstein, ``Principal warps: Thin-plate splines and the decomposition of
  deformations,'' \emph{IEEE Transactions on pattern analysis and machine
  intelligence}, vol.~11, no.~6, pp. 567--585, 1989.

\bibitem{chang2015shapenet}
A.~X. Chang, T.~Funkhouser, L.~Guibas, P.~Hanrahan, Q.~Huang, Z.~Li,
  S.~Savarese, M.~Savva, S.~Song, H.~Su \emph{et~al.}, ``Shapenet: An
  information-rich 3d model repository,'' \emph{arXiv preprint
  arXiv:1512.03012}, 2015.

\end{thebibliography}

\begin{IEEEbiography}[{\includegraphics[width=1in,height=1.25in]{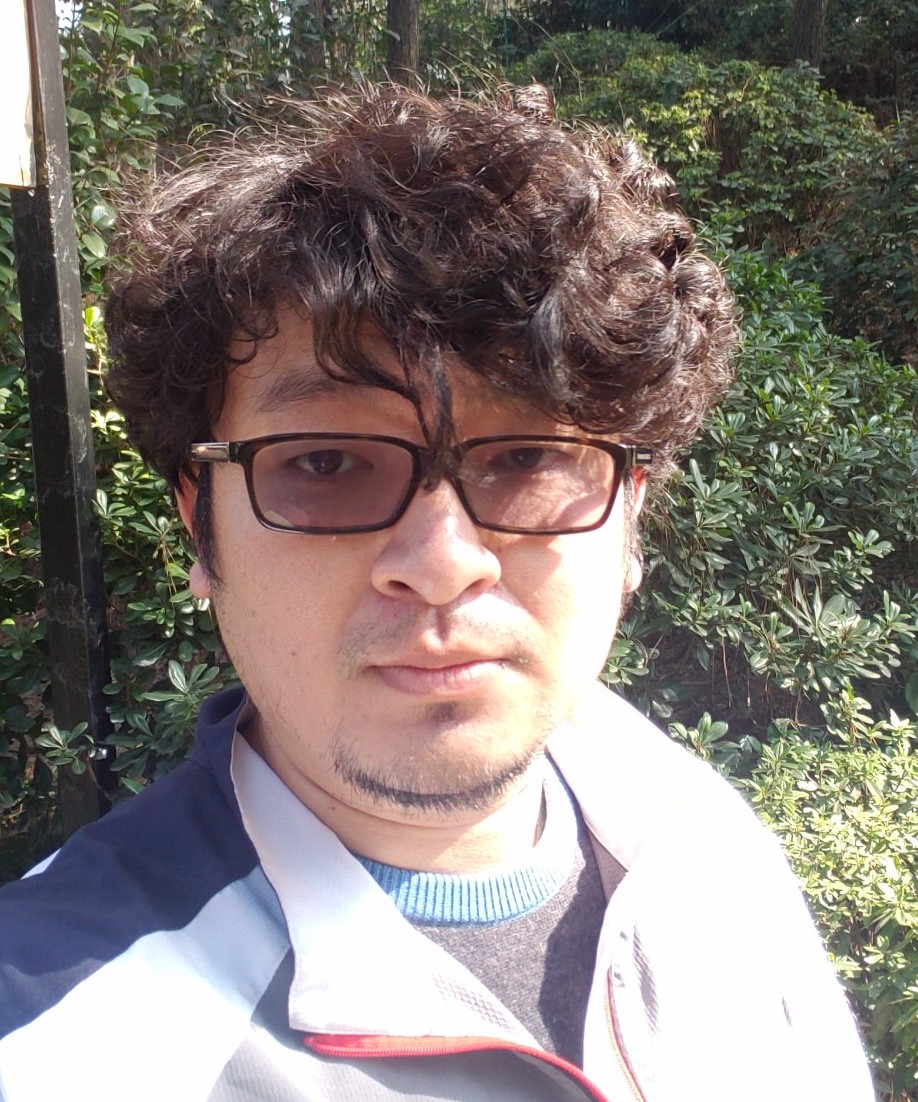}}]
{Lingjing Wang} received a B.S. degree from Moscow State University, Moscow, Russia in 2011. He received a Ph.D. degree at the Courant Institute of Mathematical Science, New York University, the USA in 2019. His research interests include deep learning and 3D visual computing.
\end{IEEEbiography}
\begin{IEEEbiography}[{\includegraphics[width=1in,height=1.25in,clip,keepaspectratio]{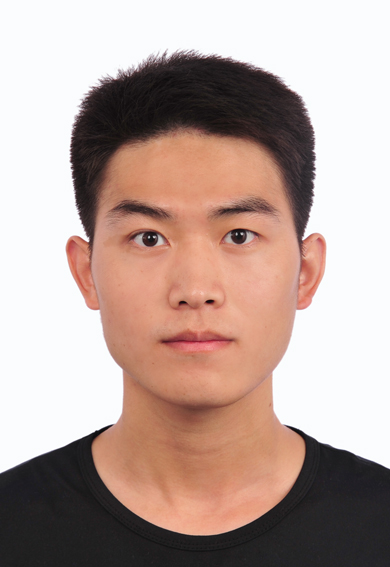}}]
{Xiang Li} received a B.S. degree in remote sensing science and technology from Wuhan University, Wuhan, China in 2014. He received a Ph.D. degree at the Institute of Remote Sensing and Digital Earth, Chinese Academy of Sciences, Beijing, China in 2019.
His research interests include deep learning, computer vision, and remote sensing image recognition.
\end{IEEEbiography}
\begin{IEEEbiography}[{\includegraphics[width=1in,height=1.25in]{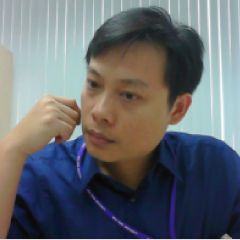}}]
{Yi Fang} received the B.S. and M.S. degrees in biomedical engineering from Xi’an Jiaotong University, Xi’an, China, in 2003 and 2006, respectively, and the Ph.D. degree in mechanical engineering from Purdue University, West Lafayette, IN, USA, in 2011.
He is currently an Assistant Professor with the Department of Electrical and Computer Engineering, New York University Abu Dhabi, Abu Dhabi, United Arab Emirates. His research interests include three- dimensional computer vision and pattern recognition, large-scale visual computing, deep visual computing, deep cross-domain and cross-modality multimedia analysis, and computational structural biology.
\end{IEEEbiography}




\end{document}